\documentclass[runningheads]{llncs}
\usepackage{graphicx}

\usepackage{amssymb}
\usepackage{amsmath}

\DeclareGraphicsExtensions{ps}
\graphicspath{{fig/}{fig1/}{tmp/}{../fig/}}
\def\mi{\begin{equation}}
\def\mf{\end{equation}}
\def\mia#1\mfa{\begin{align}#1\end{align}}
\def\miar#1\mfar{\begin{eqnarray}#1\end{eqnarray}}
\def\mmi#1\mmf{\begin{multline}#1\end{multline}}
\def\<#1\>{\begin{equation}#1\end{equation}}
\def\[#1\]{\begin{equation}#1\end{equation}} 

\def\med{\overline}

\def\reff#1{(\ref{#1})}

\newcommand{\ud}{\mathrm{d}}

\renewcommand{\v}[1]{\ensuremath{\mathbf{#1}}} 
\newcommand{\gv}[1]{\ensuremath{\mbox{\boldmath$ #1 $}}}  
\let\baraccent=\= 
\renewcommand{\=}[1]{\stackrel{#1}{=}} 

\newcommand{\transp}{\top}

\begin{document}
\title{Kernel embedded nonlinear observational mappings in the variational mapping particle filter}
\author{Manuel Pulido\inst{1,2}, Peter Jan vanLeeuwen \inst{1,3} and Derek J. Posselt \inst{4}}
\authorrunning{M. Pulido et al.}
\institute{Department of Meterology, University of Reading, UK \and
 Department of Physics, Universidad Nacional del Nordeste, Argentina \and
 Department of Atmospheric Science, Colorado State University, USA \and
 Jet Propulsion Laboratory, California Institute of Technology, Pasadena, CA, USA}

\titlerunning{Nonlinear observational mappings in the variational mapping particle filter}

\maketitle
\begin{abstract}
Recently, some works have suggested  methods to combine variational probabilistic inference with Monte Carlo sampling. One promising approach is via local optimal transport. In this approach, a gradient steepest descent method based on local optimal transport principles is formulated to transform deterministically point samples from an intermediate density to a posterior density.  The local mappings that transform the intermediate densities are embedded in a reproducing kernel Hilbert space (RKHS). This variational mapping method requires the evaluation of the log-posterior density gradient and therefore the adjoint of the observational operator. In this work, we evaluate nonlinear observational mappings in the variational mapping method using two approximations that avoid the adjoint, an ensemble based approximation in which the gradient is approximated by the particle covariances in the state and observational spaces the so-called ensemble space and an RKHS approximation in which the observational mapping is embedded in an RKHS and the gradient is derived there. The approximations are evaluated for highly nonlinear observational operators and in a low-dimensional chaotic dynamical system. The RKHS approximation is shown to be highly successful and superior to the ensemble approximation.

\end{abstract}

  \section{Introduction}
There is a large number of applications in which the process of interest is not directly measured, a latent process,  but it is related through a map to another process which is the one observed. This is a classical inverse problem, in which the latent process is infered from indirect observations \cite{tarantola05}. The mapping between the two processes will here be referred to as the observational mapping.  Depending on the application, the observational mapping may be (partially) known through the knowledge of the physical processes involved.  An example of particular interest in this work is the inference of atmospheric state variables from satellite measurements of radiation.  In other applications, the map is unknown and needs to be estimated. This is one of the central aims in machine learning applications \cite{vapnik13}.

In modeling and predicting the atmosphere, clouds play a central role. Measurements from spaceborne radars may give information on cloud properties. In this case, the observed variables are radar reflectivity and microwave radiances, while the variables of interest are cloud particle concentrations and distributions. This mapping is represented in models through parameterizations which relate cloud microphysical processes to precipitation and radiative fluxes. In several situations, the joint posterior density of model parameters and the output variables is bimodal \cite{posselt10}. The main factor responsible for the bimodal density is the extremely nonlinear response of model output variables to changes in microphysical parameters. Parameter prior density and observation uncertainty only play a secondary role in the resulting complexity of the posterior density.

If the latent process is governed by a time evolving stochastic dynamical system, the inference is sequential. The time evolution of the latent state is given by a Markov process, while an observational mapping relates observations with the latent state. These are known as state-space models or hidden Markov models. A rather general method for inference in hidden Markov models is based on Monte Carlo sampling of the prior density, referred as sequential Monte Carlo or particle filtering \cite{gordon93}. One of the major challenges in high-dimensional particle filtering is to concentrate sample points in the typical set so that they produce a non-negligible contribution to expectation estimations. 
Therefore, sample points are required to be located in the typical set to make the most of them.

Recent works propose to combine variational inference with Monte Carlo sampling \cite{saeedi17}. A rigorous well-grounded framework to combine them is via local optimal transport \cite{marzouk17,pulido18}. Optimal transport relates a given density with a target density trough a mapping which minimizes a risk. Hence, optimal transport concepts may be used to move sample points where they maximize the amount of Shannon information  they can provide. If the mapping function space is constrained to a reproducing kernel Hilbert space, the local direction that minimizes the risk, in terms of the Kullback-Leibler divergence, is well defined. This direction minimizes the Stein discrepancy \cite{liu16}. An application of the variational mapping using the Stein gradient to sequential Monte Carlo methods in the framework of hidden Markov models  was recently developed in \cite{pulido18} referred to as variational mapping particle filter (VMPF).

The gradient of the observation likelihood depends on the adjoint of the observational mapping. Thus, most of the approximations used for posterior inference  including MAP estimation, the Kalman filter, the stochastic and square-root ensemble Kalman filters (e.g. \cite{burgers98,hunt07})  require this adjoint of the observational mapping. However, there is a rather large number of complex observational mappings for which the adjoint is not available. In the context of the ensemble Kalman filter, an ensemble approximation of the adjoint of the observational mapping is used \cite{houtekamer01,hunt07}. However, this approximation may have a detrimental effect in the inference for the highly nonlinear observational mapping of e.g. cloud parameter estimation \cite{posselt12,posselt14,posselt16} and in other geophysical applications \cite{evensen18}.



A description of the the VMPF in the context of inverse modeling is given in Section \ref{sec:invm}. Two approximations of the adjoint of the observational mapping based on sample points evaluations of the observational operator are derived in Sections \ref{sec:Hrkhs} and \ref{sec:Hens}). 
Details of the experiments are given in Section \ref{sec:exp}. The VMPF with the exact gradient of the logarithm of the posterior density and the developed approximations is evaluated with nonlinear observational operators in low-dimensional spaces (Section \ref{sec:results}). The performance of the VMPF in a chaotic dynamical system with a nonlinear observational mapping is also discussed.

\section{Inverse modeling with variational mappings} \label{sec:invm}

Suppose we want to determine a stochastic latent process $\v x$ in $\mathbb{R}^{N_x}$, however data are from another process $\v y$ in $\mathbb{R}^{N_y}$. The relationship between the processes is given through a known nonlinear observational operator $\mathcal H$ such that
\mi
\v y_k=\mathcal{H}(\v x_k)+ \gv \eta_k,
\mf
where $\gv \eta_k$ is the random observational error which are realizations from $p(\gv \eta)$, $k$ denotes different realizations of the stochastic process. 
We also assume the observational errors are unbiased. 

Using Bayes rule, the density of the latent process conditioned on the realizations of the observed process is
\mi
p(\v x| \v y) \propto p(\v y|\v x) p(\v x). \label{posterior-prop}
\mf
Let us assume  prior knowledge of $\v x$ is through a sample $\{\v x^j,j=1,\cdots,N_p\}\triangleq \v x^{1:N_p}$. A standard importance sampling technique assumes that the prior density $p(\v x)$ is a proposal density of $p(\v x| \v y)$ so that this posterior distribution is written as the sample points of the prior density but evaluating the likelihood of the sample points which will represent the weights of the sample points \cite{gordon93}. 

Our aim is to transform from sample points of $p(\v x)$ to sample points of $p(\v x| \v y)$ via a series of mappings, $\v x_i = T_i(\v x_{i-1})$. Considering the mappings, the relationship between the  density after the mappings and the initial density is
\mi
q(\v x_I (\v x_0)) =\prod_{i=1}^I |\nabla T_i| \, q(\v x_0),
\mf
where the initial density $q(\v x_0)$ is in principle the prior density, the target density of the transformations is the posterior density $p(\v x_0|\v y)$ 
 and $|\nabla T_i|$ are the Jacobians of the transformations.

Therefore, in order to get equally-weighted sample points that optimally represent the posterior density, we have to find a series of maps $T$ that transform the prior in the posterior density. In terms of the particles, we have to drive them from the prior density to the posterior density. In this work, sample points will also be referred as particles indistinctly. This process of driving the particles from one to other density, could be framed as maximizing the likelihood of the particles. Alternatively, it can be formulated as a Kullback-Leibler (KL) optimization given the well-known equivalence between marginal likelihood maximization and  KL minimization. The KL divergence between the intermediate density and the target density is
\mi
KL(q_T\Vert p) = \int  q_T(\v x) \log \left[\frac{q_T(\v x)}{p(\v x|\v y)}\right] \ud \v x
\mf

The aim is to determine the {\em local} transformation $T$ that produces the deepest descent in KL. The derivation of the steepest descent gradient has been already given in some previous works \cite{liu16,pulido18}. Assuming the transformation $T$ is in a reproducing kernel Hilbert space (RKHS), the gradient of the KL divergence is given by
\mi
\nabla KL(\v x) = \int \left[K(\v x',\v x) \nabla \log p(\v x'|\v y) + \nabla K(\v x', \v x)\right] \ud \v x' \label{gradKL}
\mf
where $K(\v x',\v x)$ is the reproducing kernel and the gradient is at $\v x$ where the local transformation is produced.

Each of the particles is moved along the steepest descent direction $\v v(\v x)= - \nabla KL$,
\mi
\v x^j_{i+1} = T_{i+1}(\v x_i^j)= \v x^j_i + \epsilon \v v(\v x^j_i). \label{flow}
\mf
The particles are tracers in a flow given by the KL gradient. In essence, the objective is to determine the direction of steepest descent at each sample point and to move them along these directions. The pseudo-time step $\epsilon$ in \reff{flow} should be small enough so that the particle trajectories do not intersect and therefore the smoothness of the flow is conserved.  The overall performance of the variational mapping in a sequential Monte Carlo algorithm is evaluated in \cite{pulido18} and is termed as variational mapping particle filter (VMPF).

To obtain the gradient of the Kulback-Leibler divergence at a sample point \reff{gradKL}, we require an analytical expression of the log-posterior gradient, which can be expressed in terms of the prior density and the observation likelihood using \reff{posterior-prop},
\mi
\nabla \log p(\v x| \v y) = \nabla \log p(\v x) + \nabla \log p(\v y | \v x) \label{nablalogp}
\mf

Assuming Gaussian observational errors, $p(\gv \eta)\sim \mathcal{N}(\v 0,\v R)$, and using \reff{posterior-prop}, the log observation likelihood gradient is
\mi
\nabla \log p(\v y | \v x) =  (\nabla \mathcal{H}(\v x))^\transp \v R^{-1} (\v y-\mathcal{H}(\v x)). \label{obsLik}
\mf
The observational operator has a  major role in \reff{obsLik}. For a linear observational mapping, a linear log observation likelihood gradient results. On the other hand, a nonlinear mapping produces a nonlinear likelihood gradient and therefore it induces a non-Gaussian posterior distribution. In the case of a non-injective mapping, more than one root of \reff{obsLik} are expected, which results in a multimodal posterior density. This rather common feature in the observational mapping is examined exhaustively in the experiments.

\subsection{Observational mapping in the RKHS}\label{sec:Hrkhs}


For many applications in geophysical systems, the observational operator is a rather complex mapping that involves physical processes, for instance as mentioned the inversion from radar reflectivity to cloud properties. Even when the observational operator is available, the tangent linear and the adjoint of the observational mapping are not available and their development and use could be costly in terms of human resources and computationally demanding its evaluation. In this work, we derive a Monte Carlo approximation  of the term $(\nabla \mathcal{H}(\v x))^\transp$ in \reff{obsLik}. In coherence with the formulation of the variational mapping, we now also assume that the process $\mathcal{H}(\v x)$ is in the reproducing kernel Hilbert space (RKHS). This assumption is coherent with support vector machines which indeed also assumes the mapping lies in an RKHS \cite{vapnik13}. In that case, we can use the reproducing property for $\mathcal H(\v x)$,
\mi
\mathcal H(\v x) = \int \mathcal H(\v x') K(\v x,\v x') \ud \v x'. \label{reprProp}
\mf
Using the $N_p$ particles $\v x^{1:N_p}$ to generate a finite Hilbert space, the Monte Carlo approximation of the gradient of \reff{reprProp} is
\mi
\nabla \mathcal H(\v x) \approx \frac{1}{N_p} \sum_{j=1}^{N_p} \mathcal H(\v x^j) \nabla K(\v x,\v x^j). \label{nablaH}
\mf
We have now an expression of the gradient of the observational operator that only depends on its evaluation at the  particles. From the RKHS theory, we know that the approximated  value in \reff{nablaH} will converge towards the exact one when $N_p\rightarrow \infty$ assuming $\mathcal H(\v x)$ is sufficiently smooth. Convergence of the gradient in the RKHS has been examined in \cite{zhou08}.

The expression of the gradient of the Kullback-Leibler divergence of the VMPF \reff{gradKL} using a Monte Carlo integration is
\mi
\nabla KL(\v x) = - \frac{1}{N_p} \sum_{l=1}^{N_p}  \left[ K(\v x^{l}, \v x ) \nabla \log p(\v x^{l}|\v y)  +  \nabla K(\v x^{l}, \v x)\right].\label{gradKL1}
\mf
using \reff{nablalogp} and \reff{nablaH} in \reff{gradKL1}, the gradient becomes
\begin{align}
\nabla KL(\v x) = - \frac{1}{N_p} \sum_{l=1}^{N_p} & \left\{ K(\v x^{l}, \v x ) \left[\nabla \log p(\v x^{l}) +  \left(\frac{1}{N_p} \Sigma_j \mathcal H(\v x^j) \nabla K(\v x^l,\v x^j)\right)^\transp \right. \right. \nonumber \\
    & \left.\left.  \v R^{-1} (\v y-\mathcal{H}(\v x)) \right]   +  \nabla K(\v x^{l}, \v x)\right\}.\label{gradKL3}
\end{align}

This expression depends only on the evaluation of the observational operator at the sample points. Therefore, the number of evaluations of the observational operator in \reff{gradKL3} is $N_p$ at each mapping iteration. No extra evaluations from the original variational mapping are required. 
The Gram matrix and the gradient of the kernels are already available since they are required in the second right-hand side term of \reff{gradKL1}. In conclusion, the main complexity of the algorithm is still of order $N_p^2$ as the original VMPF.

There is a problem for the RKHS approximation of the observational mapping \reff{nablaH} in the regions where the sample points are sparse. An experiment to illustrate this drawback is shown in Section 3. This problem appears because the kernel values between those sparse points and the rest of the sample points have only few points (the closest ones to the one in consideration) with non-negligible contributions and all the other kernel values are (close to) 0. Note that the square of the bandwidth is chosen to be smaller than the trace of the sample covariance to allow for more detailed structures in the density of $\mathcal H(\v x)$. One way to solve this problem could be using an adaptive kernel bandwidth based on the distance to the k-nearest neighbors. A simpler solution is to normalize the contributions of the kernels
\mi
\mathcal H(\v x) \approx \frac{\sum_{j=1}^{N_p} \mathcal H(\v x^j) K(\v x,\v x^j)}{\sum_{l=1}^{N_p} K(\v x,\v x^l)} \label{HrkhsNor}
\mf
In this way, the contribution of the kernel functions evaluated at each sample point is normalized. 
This approximation to the gradient of the observational mapping is evaluated in the experiments.

\subsection{Observational operator in the ensemble space}\label{sec:Hens}
Instead of constraining the observational operator to the RKHS,  it can be expressed in the ensemble perturbation space. This type of approximations is common in ensemble Kalman filtering. Indeed, the whole estimation problem may be transformed and determined in the ensemble perturbation space \cite{hunt07}. Here, we derive an approximate expression for the tangent-linear model, i.e. the gradient of the observational mapping, and its adjoint model based on the perturbations of the particles (ensemble members) to the mean.

The increments are approximated with a first-order Taylor series around the mean $\med{\v x}$
\mi
\v y-\mathcal{H}(\v x)\approx \v y-\mathcal{H}(\med{\v x}) - \v H (\v x - \med{\v x}),
\mf
where $\v H$ is the tangent-linear of $\mathcal{H}$ at $\med{\v x}$. Using the perturbation matrices in the state and observational spaces, whose the columns are the differences between the ensemble members and the mean, we find
\mi
\v X= \frac{1}{\sqrt{N_p-1}} \left(\v x^{(1)}-\med{\v x},\v x^{(2)}-\med{\v x}, \cdots, \v x^{(N_p)}-\med{\v x}\right), \label{Xper}
\mf
\mi
\v Y= \frac{1}{\sqrt{N_p-1}} \left(\mathcal{H}(\v x^{(1)})-\med{\mathcal{H}(\v x)},\cdots, \mathcal{H}(\v x^{(N_p)})-\med{\mathcal{H}(\v x)}\right). \label{Yper}
\mf
The included normalization factor is $\sqrt{N_p-1}$ to avoid bias in the sample covariance.

The approximated tangent linear operator of $\mathcal{H}$ at the ensemble space is then given by
\mi
\nabla \mathcal H(\overline{\v x}) = \v H\approx  \v Y \v X^{-1} = \v Y \v X^\top (\v X \v X^\top)^{-1}= \v P_{yx} \v P^{-1}, \label{H_ES}
\mf
where $\v P_{yx}\triangleq \v Y \v X^\top $. Similar approximations have been used in applications of the ensemble Kalman filter.  This ensemble-based approximation of the derivative of the observational operator is coherent with the one used in the Kalman gain \cite{houtekamer01},
\[ \v P \v H^\top = \v P (\v P_{yx} \v P^{-1})^\top  = \v X \v Y^\top. \]

For the adjoint approximation, the transpose of \reff{H_ES} is used,
\[ \nabla \mathcal{H}(\overline{\v x})^\transp \approx \v H^\top = \v X^{-\top} \v Y^\top. \]

\section{Experiments}\label{sec:exp}

In the experiments, an iid sample from the prior density is given, which for simplicity is assumed in general to be normally distributed. Furthermore, observational errors are assumed Gaussian. Thus, the only source of non-Gaussianity in the posterior density is the nonlinearity in the observational operator. The prior density is $\mathcal N(0.5,1)$.  The observation corresponds to a true state of $3$ with an observational error of $R=0.5$. In the case of the nonlinear dynamical system, the prior density in subsequent cycles is affected by both the nonlinear mapping of the dynamical system and of the observational operator, so that it is non-Gaussian.


The approximations are evaluated with two nonlinear observation operators. We use a quadratic relationship,  $\mathcal H(x) = x^2$, which  is expected to lead to  a bimodal posterior density because of its surjectivity. For a challenging evaluation of the gradient approximations of the observational operator, we also use the absolute value $y=|x|$ which contains a discontinuity in its derivative. Representations of the observational mapping through a small number of basis functions is expected to give an inaccurate approximation of this derivative. Although these observational operators are only motivated in evaluating the approximations, they are indeed found in several applications. In particular, the absolute value is a frequent operator that appears when measuring wind and current speeds with instruments that are not able to distinguish flow direction.

Note that the prior density we use is not symmetric around 0, while the chosen observational operators are. Besides, the true state is in a region of very small prior density. These choices have been taken so that the gradient approximation from sample points represents a challenge.

Another set of experiments assume a uniform prior density. These experiments are motivated in the densities of physical parameters, in particular related to cloud parameterizations, which may be physically constrained between two extreme values while there is no apriori knowledge of the parameter values inside the domain. To account for these densities in the variational mapping, we introduce a hard boundary to the particles at the parameter extreme values. Particles that reach the boundaries are reflected as in a wall. We allow that part of the attractor, e.g. one of the modes of the posterior density, may be outside the domain and therefore the particles will tend in that case to remain asymptotically at the boundaries.

In a last set of experiments, we evaluate the use of a nonlinear observation operator in a chaotic dynamical system. The state of the 3-variable Lorenz-63 dynamical system corresponds to the latent process. Observations are obtained with the absolute  observational mapping from the latent state and a Gaussian noise. The results from the VMPF using 100 particles are compared with the SIR particle filter \cite{gordon93} using 1000 and 10000 particles.

The optimization in the VMPF is conducted through ADAM \cite{kingma15}, a second-moment method. The tuning parameters are set to the recommended values, first moment parameter $\beta_1=0.9$ and second moment parameter $\beta_2=0.99$. The learning rate was set to 0.03.

\vspace{-0.5cm}
\begin{figure}
  \includegraphics[width=4.8in]{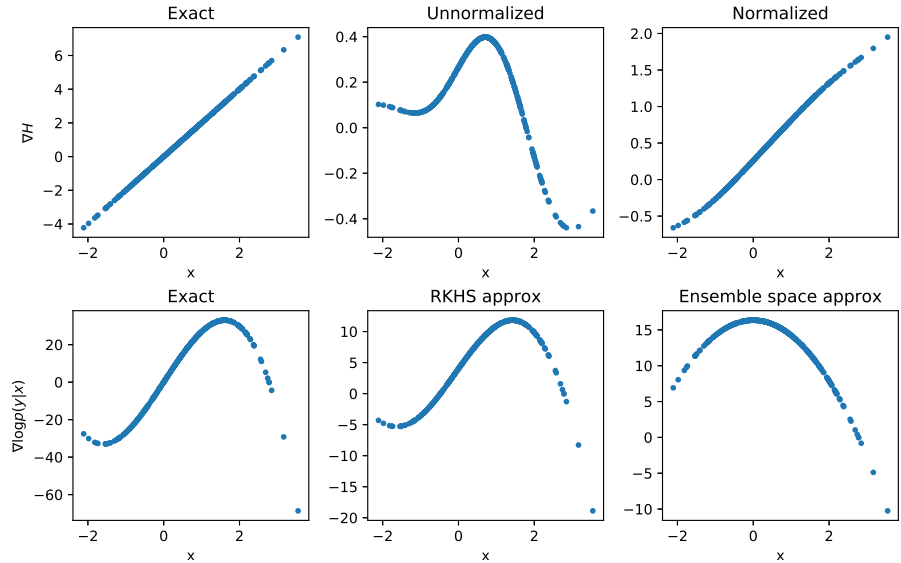}
\caption{The gradient of a quadratic observational operator, $\mathcal{H}(x)=x^2$, represented by the samples of the prior density  for the  exact  calculation (left upper panel), the approximation using unnormalized kernels (middle upper panel) and using normalized kernels (right upper panel). The gradient of the log-observation likelihood, \reff{obsLik}   with exact gradient (left lower panel), normalized RKHS approximation (middle lower panel) and ensemble approximation (right lower panel). }\label{quaH}
\end{figure}
\vspace{-1cm}


\section{Results}\label{sec:results}

Figure \ref{quaH} shows the results of the derivative of the quadratic observational operator (upper panels) represented by using the sample points of the prior density. The exact calculation is shown in left upper panel of Figure \ref{quaH}. The approximation in the RHKS using unnormalized kernel functions in the finite space, \reff{nablaH},  is in the middle panel and the one using normalized kernels, \reff{HrkhsNor}, in the right upper panel. The approximation to the mapping gives small values in isolated sample points because of only a few points contribute to the kernel integration in sparse areas. The normalization factor incorporates weights  according to the density of points around the samples, producing a better estimate of the functional representation of the mapping and its derivative. However, note that some smoothing is still found in the extremes which results in approximated derivative values smaller that the true ones. This effect in the sparse sample points should manifest in strong convex functions as the one used in the evaluation. The normalized kernel approximation--apart from the amplitude--  gives a rather good functional dependence. There are some minor deviations in the functional dependence mainly produced by the asymmetry introduced between the sampling and the observational mapping. 

The impact of approximating $\nabla \mathcal H$ on the gradient of the observation likelihood is shown in the lower panels of Fig. \ref{quaH}. The overall structure using the RKHS approximation is recovered. On the other hand, the amplitude of the gradient is underestimated. The ensemble space approximation gives a constant gradient of the observational mapping independent of the sample points, \reff{H_ES}, which is expected to give the mean gradient of the mapping. In terms of the gradient of the observation likelihood, it results in a quadratic function, because of the increment term in \reff{obsLik} between observations and the particles.  This would only be a good approximation of the true observation likelihood gradient (left panel) close to the observation. For methods that only give the maximum a posteriori solution, a relatively coarse representation of the gradient of the log-likelihood function may be enough to give a good estimation. Thus, they only require a precise gradient of the observational operator close to the observation. On the other hand, an accurate representation in a larger region is required if the inference problem also deals with the uncertainty quantification.

\vspace{-0.5cm}
\begin{figure}
\includegraphics[width=4.8in]{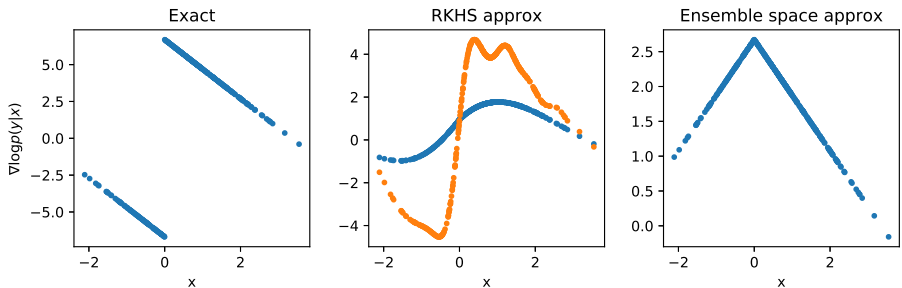}
  \vskip -.2cm
\caption{Gradient of the observation mapping $\mathcal{H}(x)=|x|$ for the exact calculation, the approximation using kernels and the approximation using perturbations in the ensemble space (upper panels). The gradient of the observation likelihood is shown in the lower panels. Two kernel bandwidths, $\gamma=1$ (blue dots) and $\gamma=0.3$ (orange dots) are shown for the RKHS approximation.}\label{absNablaH}
\end{figure}
\vspace{-0.7cm}

Results for the absolute  observational operator are shown in Fig. \ref{absNablaH}. Gaussian kernels act as smoothers (e.g. \cite{vapnik13}), so that the approximation with Gaussian kernels to the absolute function is a smooth function and so the derivative is similar to a tanh-function with a smooth transition between the positive and negative values. The transition can be more abrupt if the kernel bandwidth is reduced from $\gamma=1$ to 0.3 (middle panel in Fig. \ref{absNablaH}). However, the sampling noise is increased in that case. Note also that the amplitude of the function approximation is closer to the true one for the narrower kernel bandwidth. A narrower kernel bandwidth uses less sample points to approximate the mapping. Hence, it diminishes the smoothing. The ensemble space average produces a correct gradient of the log-likelihood close to the observations in the positive state values, but a wrong one for negative state values (lower right panel). Because the amplitude in $\nabla \mathcal H$ results from an average of all the sample points, it is underestimated in the absolute mapping and so in the gradient of the log-likelihood.

\vspace{-0.5cm}
\begin{figure}
\includegraphics[width=1.58in]{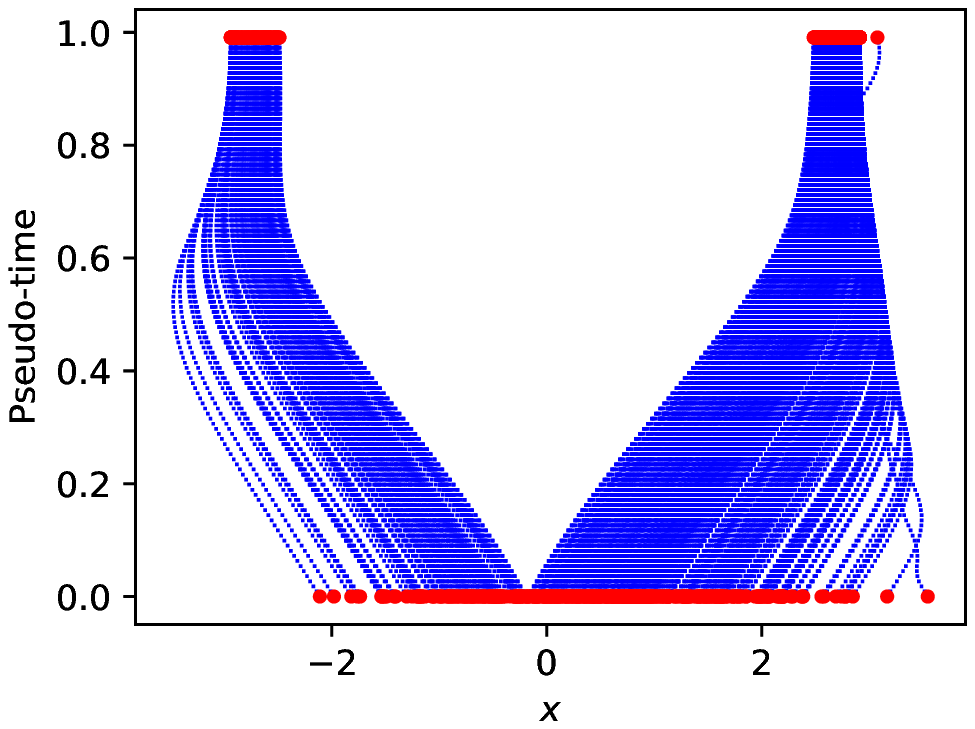}
\includegraphics[width=1.58in]{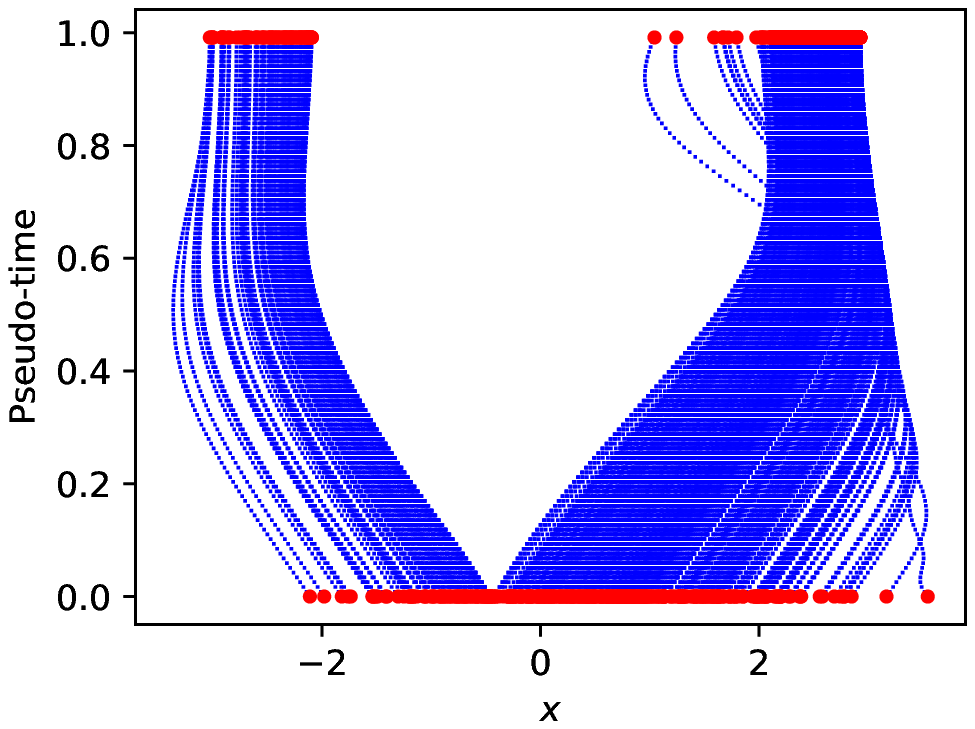}
\includegraphics[width=1.58in]{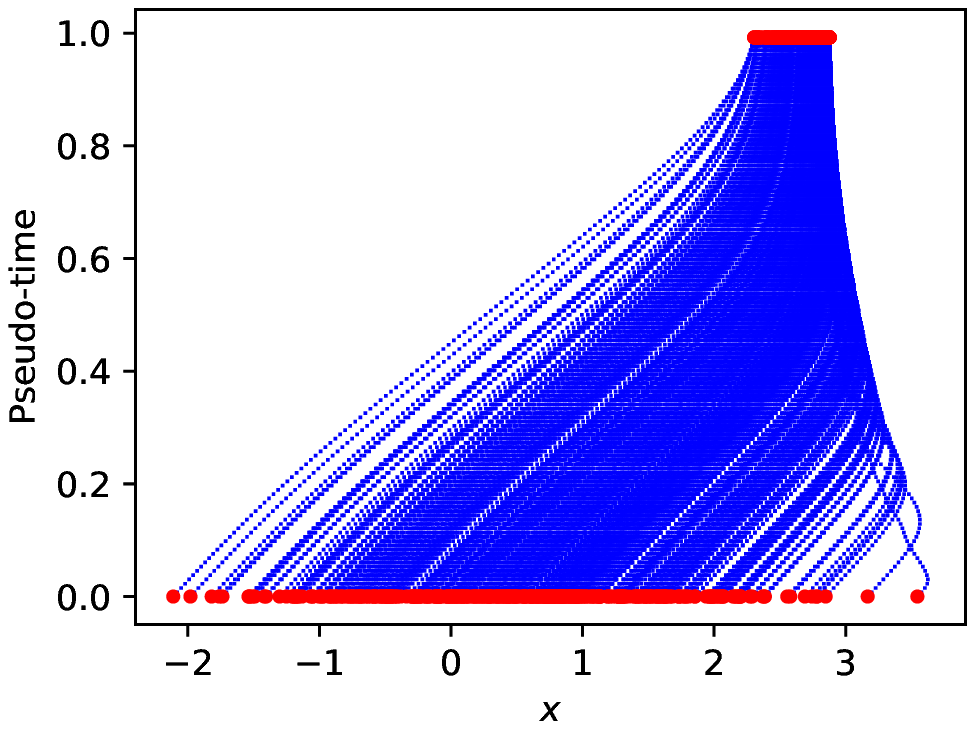}

\includegraphics[width=1.58in]{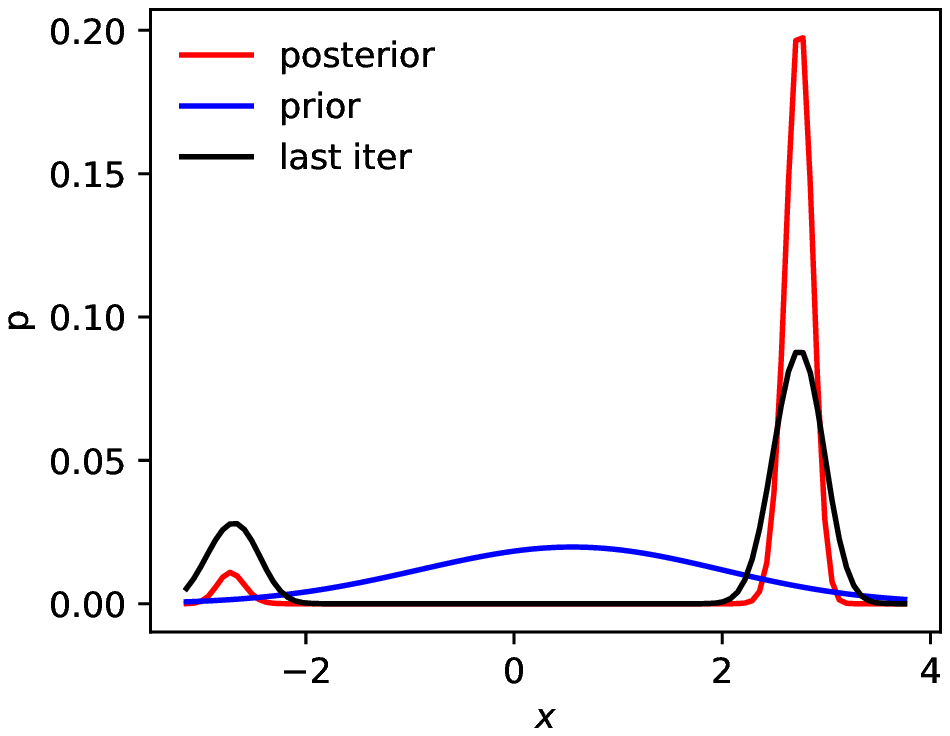}
\includegraphics[width=1.58in]{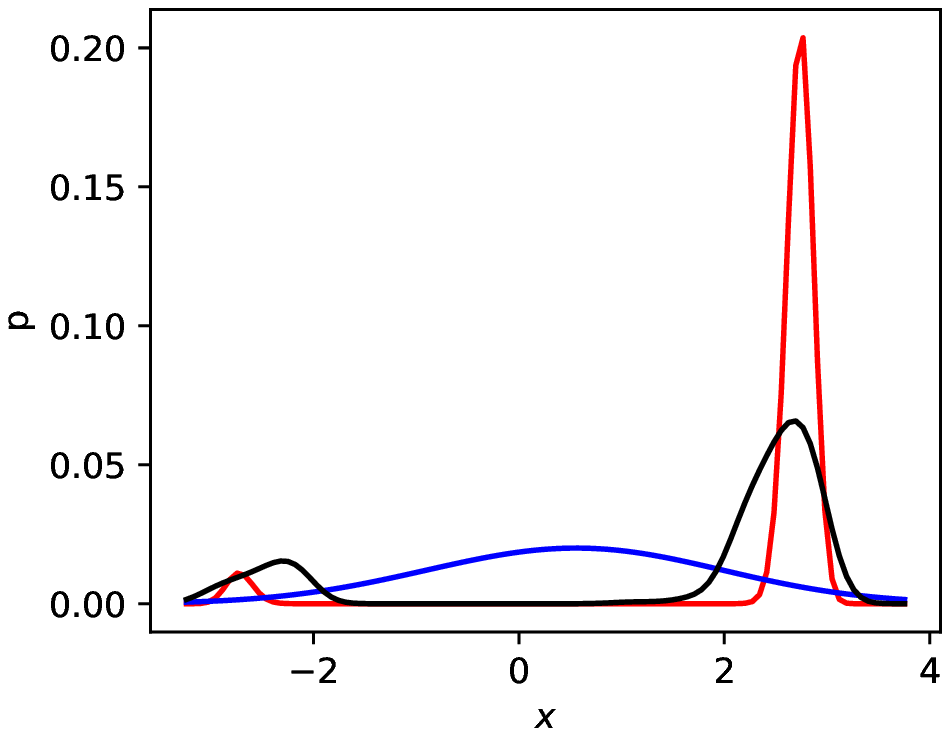}
\includegraphics[width=1.58in]{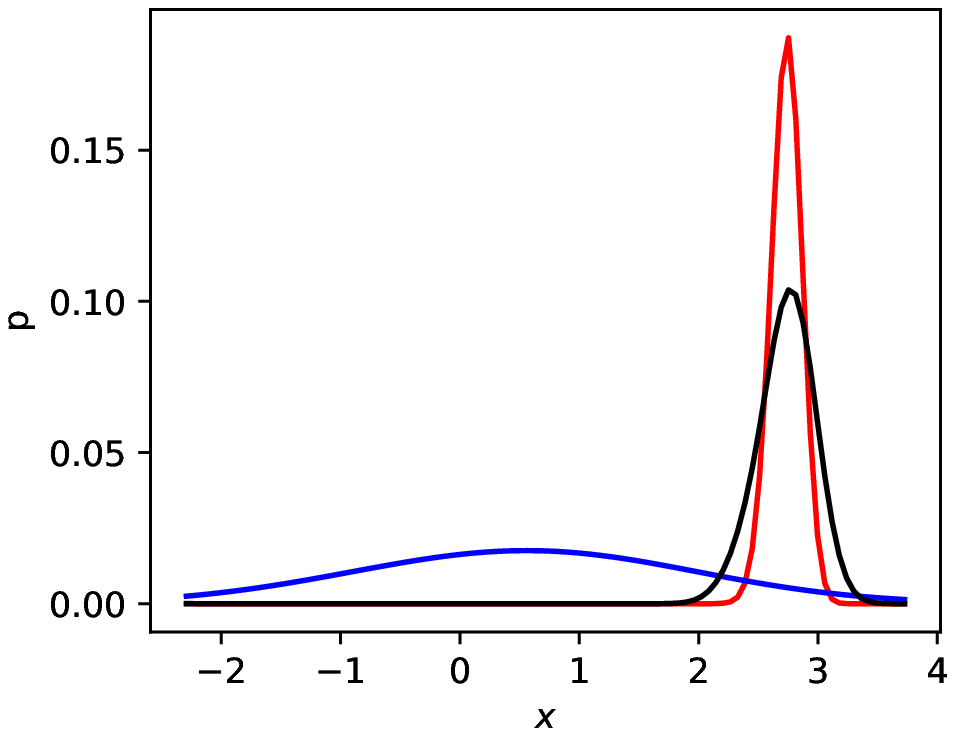}
\vskip -0.5cm
\caption{Evolution in pseudo-time of the sample points for a quadratic observational mapping (upper panels) for the experiment with exact gradient (left panel), RKHS approximation (middle panel) and ensemble approximation (right panel). Posterior density (lower panels) for the exact quadratic  observational mapping (red line) and the one obtained with VMPF (black line)}\label{dots_sqr}
\end{figure}
\vspace{-0.7cm}

The variational mapping optimizes the sample points starting from the points of the prior density. The amplitude and direction of the deterministic mappings for each particle depends on the observational mapping. The impact of the different approximations in the gradient of the observational likelihood  affects not only the rate of convergence of the VMPF but also the interactions between the particles and therefore the final distributions of the sample points that represent the posterior density. Figure \ref{dots_sqr} exhibits the trajectories of the sample points as a function of pseudo-time between the initial iteration of the filter (representing the prior density) and up to the convergence criterion is met which is based on the module of the gradient of the Kullback-Leibler divergence. The experiment corresponds to the quadratic mapping. In both the exact and the RKHS approximation, samples are attracted to two different positive/negative regions which represent a bimodal posterior density. Because of the asymmetry in the prior density (whose the mean is 0.5) more particles are attracted to the positive region. The particles finish more disperse in the RKHS approximation than in the exact gradient calculation. The ensemble space approximation for the gradient of the observational mapping removes the bimodality of the posterior density and the particles are only attracted by the dominant mode (left panel). Lower panels in Fig. \ref{dots_sqr} compare the analytical posterior density with the one obtained with the VMPF, representing the final sample with kernel density estimation in coherence with the RKHS used in the mappings. The VMPF using the exact observational mapping is shown in the left panel of Fig. \ref{dots_sqr}, the one using the RKHS approximation (middle panel) and the ensemble approximation (right panel). The exact case shows some smoothing of the main mode mainly because the observation is at a low density region of the likelihood. Tests with a narrower kernel bandwidth diminish the effect but it does not disappear. In the case of the RKHS approximation, there is some spread of the sample points toward lower values. This effect may be linked to the lower values of the gradient of the likelihood in this approximation. The ensemble approximation removes the smaller mode and only represents the main one. The slightly wider representation of uncertainty around the main mode is mainly controlled by the kernel bandwidth.

\vspace{-0.5cm}
\begin{figure}
  \includegraphics[width=1.57in]{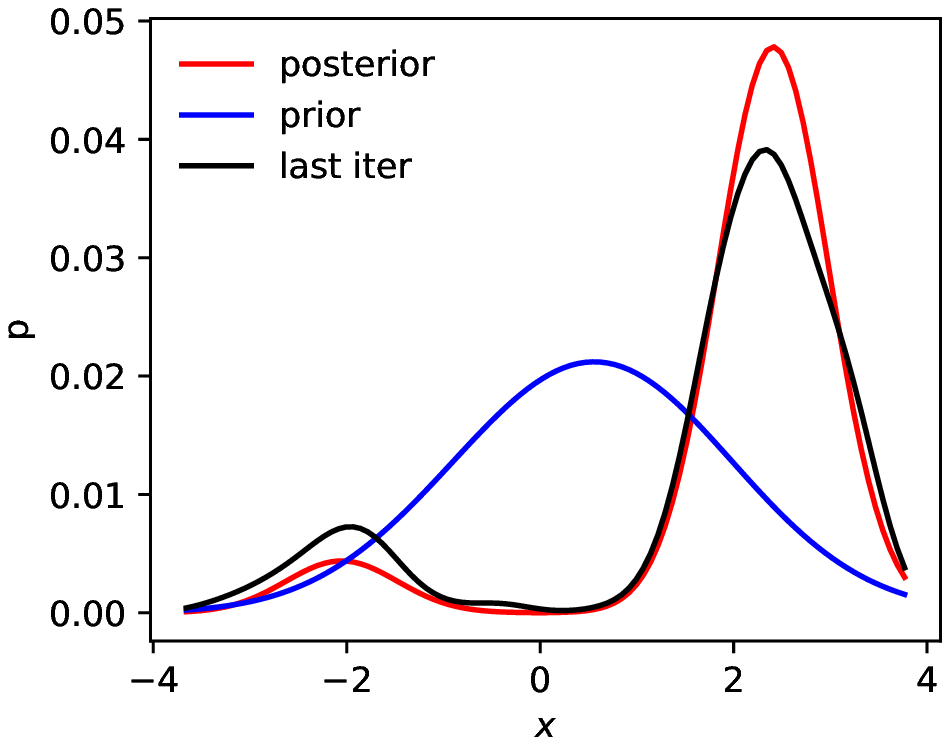}
\includegraphics[width=1.57in]{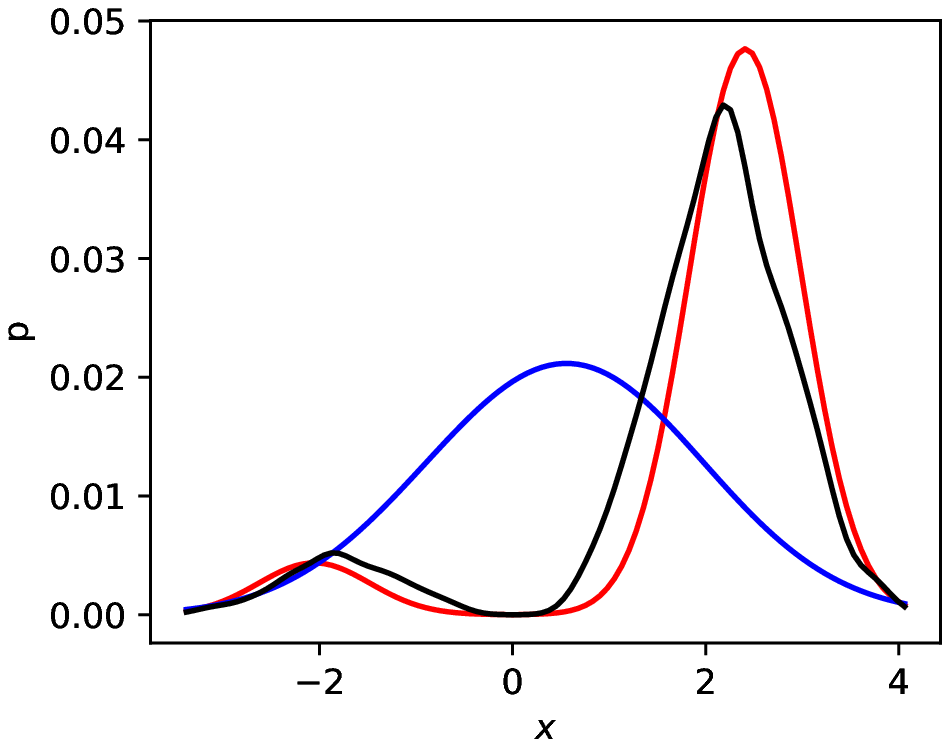}
\includegraphics[width=1.57in]{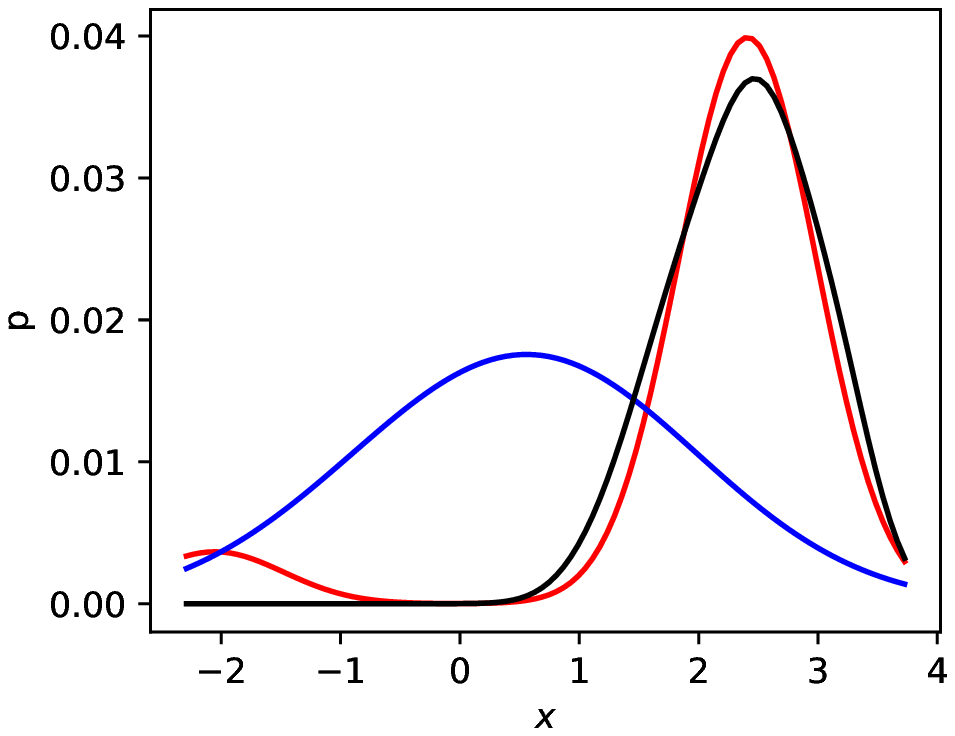}
 \vskip -0.5cm
\caption{Posterior density for the absolute  observational mapping. Panels as in Fig. \ref{dots_sqr}.}\label{post_abs}
\end{figure}
\vspace{-0.5cm}

The posterior densities resulting from the absolute observational mapping experiment are shown in Fig. \ref{post_abs}. Because of the piece-wise linear mapping, the final posterior density uncertainty is comparable with the prior density. Furthermore, the mode of the posterior density is larger than for the quadratic mapping and therefore both modes are better represented in the variational mapping particle filter with some relatively (compared to the previous case) smaller smoothing of the main mode. The RKHS approximation makes a very good job, indeed the amplitude of the modes appear to be better represented in this case. However, it is still evident a small shifting of both modes towards smaller state values. The ensemble approximation only represents the main mode. The uncertainty of this mode is well reproduced but with a slight overestimation.

\begin{figure}
  \includegraphics[width=1.58in]{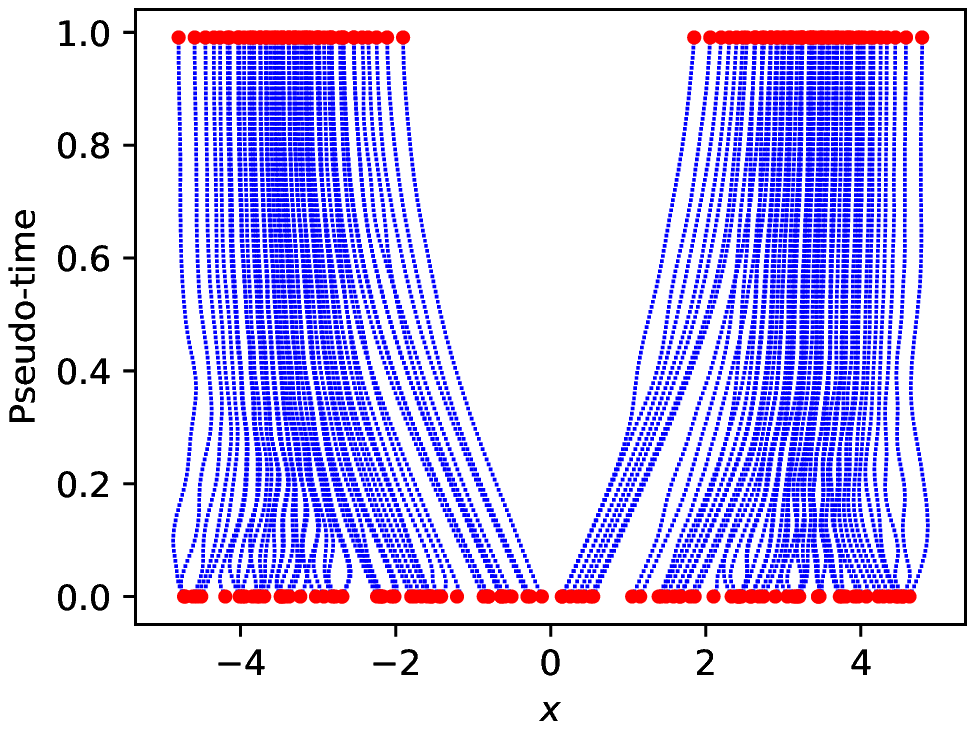}
\includegraphics[width=1.58in]{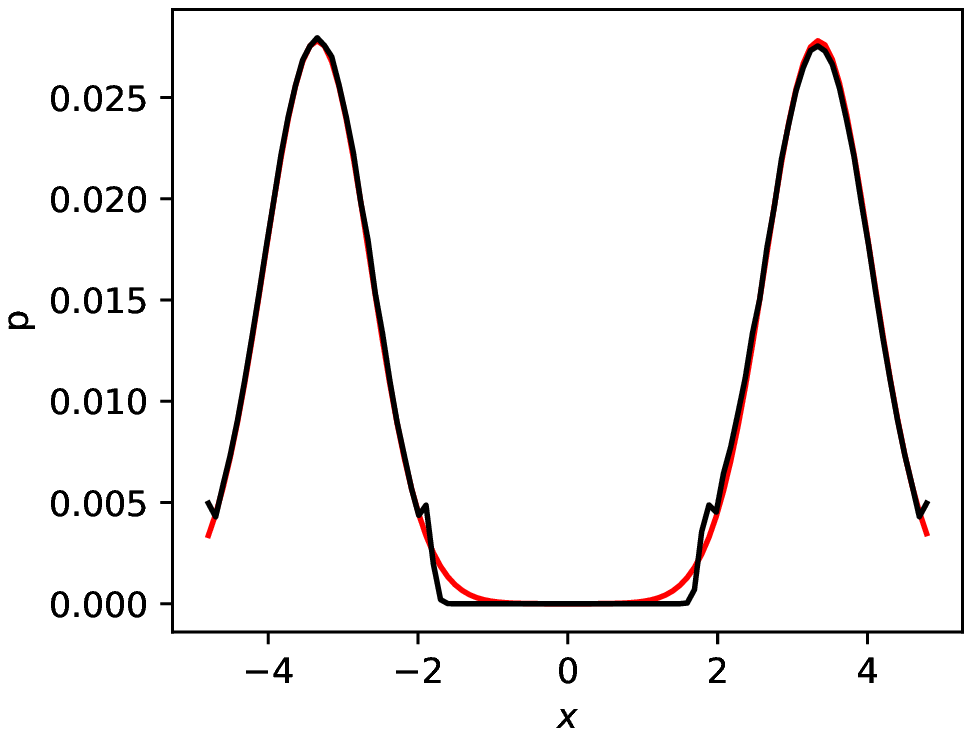}
\includegraphics[width=1.58in]{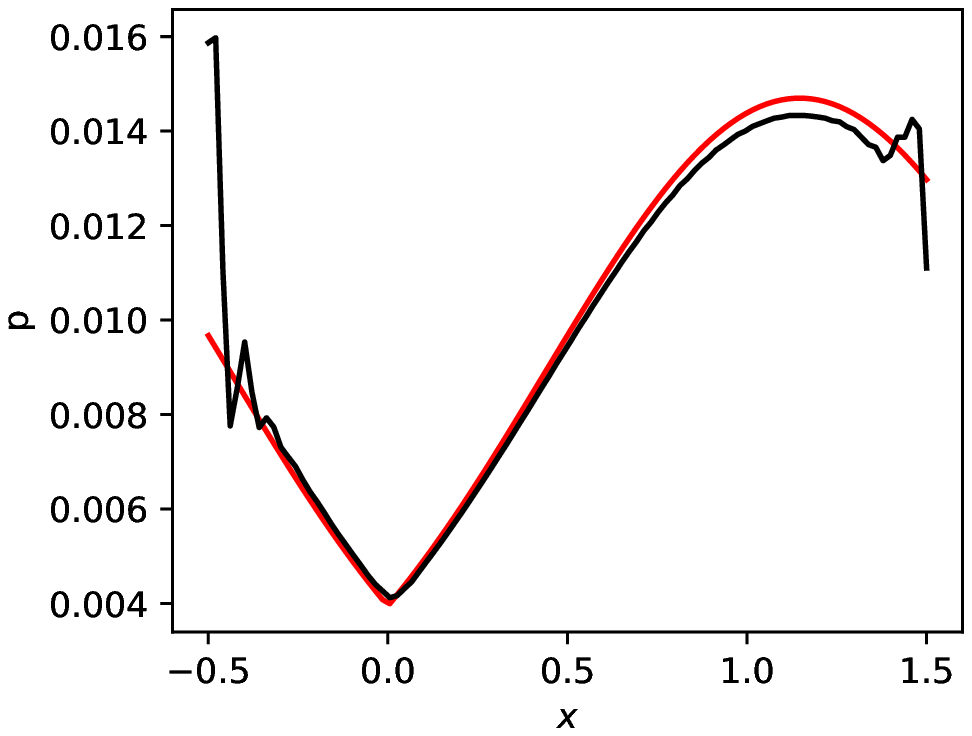}
 \vskip -0.5cm
\caption{Evolution of particle trajectories for  the absolute  observational mapping with a uniform prior density using a domain with well-captured modes (left panel), the corresponding posterior density (middle panel) and the posterior density with a mode outside the domain and large discontinuities in the posterior density (right panel). Red lines shows the analytical posterior distribution and black line the posterior density obtained with the variational mapping. }\label{post_uni}
\end{figure}

Figure \ref{post_uni} (left panel) shows the pseudo-time evolution of the particles using a  uniform prior density $\mathcal U(-5,5)$ and the  absolute  observational mapping with the true state at 3.  The resulting analytical posterior density is compared with the result obtained with the variational mapping in the middle panel of Fig. \ref{post_uni}. An overall excellent agreement between the densities is found, the technique captures the bimodal structure very well. A small edge effect is found at the boundary. Besides, an underestimation at the tails in  small values is also visible. For  an evaluation of edge effects, a uniform prior density   $\mathcal U(-0.5,1.5)$ and true state at 0.8 is shown in the left panel. In this case one mode is outside of the domain while  part of the high-density region around the other mode is also outside of the domain so that the posterior density exhibits two discontinuities at the boundaries. In this case an overestimation of the density close to the edges is found, which can slightly affect the overall amplitudes of the density. However, apart from the edge effects, the Monte Carlo technique produces a remarkably good approximation.

\vspace{-0.5cm}
\vspace{-0.7cm}

\begin{figure}
  \includegraphics[width=4.8in]{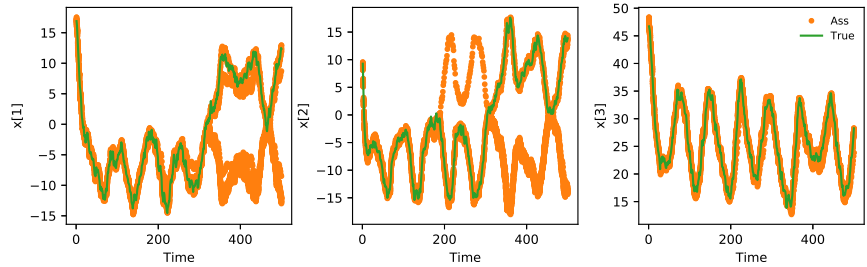}
 \vskip -0.5cm
\caption{The evolution of the latent true state variables of the stochastic Lorenz-63 dynamical system (green line) and a number of trajectories (40) of the particles resulting from the VMPF. Panels show each variable of the Lorenz 63 system.}\label{traj-l63}
\end{figure}

\begin{figure}
  \includegraphics[width=4.8in]{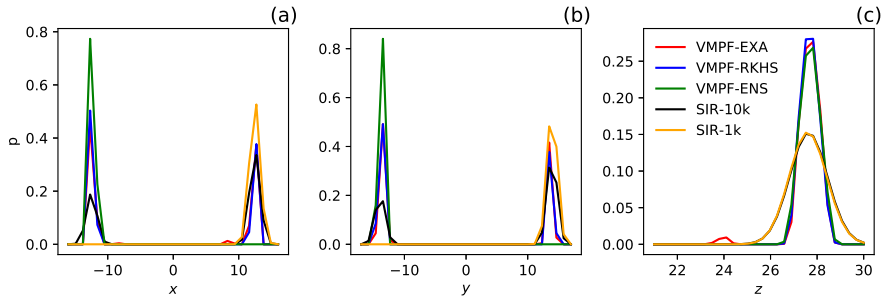}
\caption{Marginalized sequential posterior density  represented through kernel density estimation, obtained with the VMPF using the exact (VPMF-EXA), RKHS (VMPF-RKHS) and ensemble calculations (VMPF-ENS) of the adjoint. The densities from SIR particle filter with 1000 (SIR-1k) and 10000 particles (SIR-10k) are also shown. Panels show marginalized density as a function of each variable for the Lorenz 63 system at the 500 cycle.}\label{post-l63mpf}
\end{figure}

Figure \ref{traj-l63} shows the evolution of the variables of the Lorenz-63 system for a selected set of particles from VMPF. Because the apriori density is unimodal the posterior density evolves as unimodal until a transition in the true state occurs. From that time, because the absolute observations cannot distinguish in which attractor is the system, the subsequent evolution undergoes a transition to a bimodal density. Figure \ref{post-l63mpf} shows the marginal posterior densities in each variable for the VMPF. Both the exact calculation and the RKHS approximation in the observation likelihood gradient  are able to capture the bimodality of the posterior density using 100 particles. On the other hand, the ensemble approximation only gives an unimodal density. 
For comparison we also show in Fig. \ref{post-l63mpf} the corresponding marginalized posterior density of the SIR particle filter with 1000 and 10,000 particles. The SIR filter requires 10,000 particles to capture the bimodal structure of the posterior density while VMPF only requires 100 particles.  

\vspace{-0.5cm}

\section{Conclusions}

This work evaluates the use of a nonlinear observational operator in the variational mapping particle filter. The evaluation is carried out offline, as in an inverse problem with a latent process, and also online in a hidden Markov model with the time evolution of the hidden state governed by a nonlinear dynamical system. Non-injectivity of the observational mappings leads to multimodal posterior densities which is known to represent a challenge for inference methods. The variational mapping particle filter is able to capture the bimodality in the density in offline and online experiments. Particles are attracted to  the modes in coherence with the gradient of the posterior density and local optimal transport principles. In general about 50 iterations in the mapping are required to get the convergence threshold in the module of the gradient. However, for some specific cycles 200 iterations may not be enough. A small number of trajectory intersections is found, but these do not appear to affect the convergence of the overall optimization.

Two approximations of the gradient of the observation mapping are evaluated. The representation of the observational mapping in the RKHS which overall exhibits a good performance in non-Gaussian densities. Because of the smoothing associated with this representation, it may slightly shift the modes in multi-modal densities for cases in which the observation is in low-density regions of the prior density. The evaluation with the Lorenz-63 shows that the impact of this smoothing in a sequential scheme is negligible even for the absolute value observational mapping--a discontinuous gradient. The ensemble approximation of the gradient as expected does not capture multimodality, but it gives a good approximation of the main mode of the posterior density and its uncertainty. 

We have not considered here other approximations which could require further evaluations of the observational mapping apart from the sample points to estimate the gradient of the mapping. One of these possibilities is the evaluation of the gradient at each sample point from finite differences. For applications of moderate dimensions and complex observational mapping the computational cost of these further evaluations required at each iteration of the variational inference algorithm and at each sample is prohibitive.

An evaluation of a uniform prior density is conducted, which is motivated in cases of physical parameters which can only vary in a limited range (outside of this range the observational operator may give no solution or completely spurious ones). A reflexion mechanism of the particles at the boundaries is modeled in the variational mapping particle filter which accounts for these constrained optimization problems very efficiently. The constrained domain is empirically found to improve the representation of the density for a limited number of particles.


\begin{thebibliography}{8} 
  {\scriptsize
  

\bibitem{burgers98}
Burgers, G., Jan van Leeuwen, P. and Evensen, G.: Analysis scheme in the ensemble Kalman filter. {\em Monthly weather review}, {\bf 126}, 1719-1724 (1998).

\bibitem{daum07}
Daum, F. and Huang, J.: Nonlinear filters with log-homotopy. {\em In Signal and Data Processing of Small Targets 2007}. {\bf 6699}, p. 669918 (2007). 



\bibitem{evensen18}
Evensen, G.: Analysis of iterative ensemble smoothers for solving inverse problems. {\em Computational Geosciences}, {\bf 22}, 885-908 (2018).

\bibitem{gordon93}
Gordon, N.J., Salmond, D.J. and Smith, A.F.: Novel approach to nonlinear/non-Gaussian Bayesian state estimation. In {\em IEE Proceedings F-radar and signal processing}, {\bf 140},  107-113 (1993). 

\bibitem{hoffman13}
Hoffman, M.D., Blei, D.M., Wang, C. and Paisley, J.: Stochastic variational inference. {\em The Journal of Machine Learning Research}, {\bf 14}, 1303-1347 (2013).

\bibitem{houtekamer01}
Houtekamer, P.L. and Mitchell, H.L.: A sequential ensemble Kalman filter for atmospheric data assimilation. {\em Monthly Weather Review}, {\bf 129}, 123-137 (2001).

\bibitem{hunt07}
Hunt, B.R., Kostelich, E.J. and Szunyogh, I.: Efficient data assimilation for spatiotemporal chaos: A local ensemble transform Kalman filter. {\em Physica D}, {\bf 230}, 112-126 (2007).

\bibitem{jordan99}
Jordan, M.I., Ghahramani, Z., Jaakkola, T.S. and Saul, L.K.: An introduction to variational methods for graphical models. {\em Machine learning}, {\bf 37}, 183-233 (1999).

\bibitem{kingma15}
Kingma, D. and Ba, J.: Adam: A method for stochastic optimization.  {\em In Int. Conf. on Learning Repres. (ICLR)} arXiv preprint arXiv:1412.6980 (2015).

\bibitem{liu16}
Liu, Q. and Wang, D.: Stein variational gradient descent: A general purpose bayesian inference algorithm. {\em In Advances In Neural Information Processing Systems}, 2378-2386 (2016).

\bibitem{marzouk17}
Marzouk Y, T Moselhy, M Parno, A Spantini (2017) An introduction to sampling via measure transport. To appear in {\em Handbook of Uncertainty Quantification}; R. Ghanem, D. Higdon, and H. Owhadi, editors; Springer. arXiv:1602.05023 

\bibitem{posselt10}
Posselt, D.J. and Vukicevic, T.: Robust characterization of model physics uncertainty for simulations of deep moist convection. {\em Monthly Weather Review}, {\bf 138}, 1513-1535 (2010).

\bibitem{posselt12}
Posselt, D.J. and Bishop, C.H.: Nonlinear parameter estimation: Comparison of an ensemble Kalman smoother with a Markov chain Monte Carlo algorithm. {\em Monthly Weather Review}, {\bf 140}, 1957-1974 (2012).

\bibitem{posselt14}
Posselt, D. J., D. Hodyss, and C. H. Bishop: Errors in Ensemble Kalman Smoother Estimates of Cloud Microphysical Parameters, {\em Mon. Wea. Rev.}, {\bf 142}, 1631-1654 (2014).

\bibitem{posselt16}
Posselt, D. J.: A Bayesian Examination of Deep Convective Squall Line Sensitivity to Changes in Cloud Microphysical Parameters. {\em J. Atmos. Sci.}, {\bf 73}, 637-665 (2016).

\bibitem{pulido18}
Pulido M., and P. J. vanLeeuwen: Kernel embedding of maps for Bayesian inference: The variational mapping particle filter. Submitted. https://arxiv.org/pdf/1805.11380 (2018).

\bibitem{saeedi17}
Saeedi, A., Kulkarni, T.D., Mansinghka, V.K. and Gershman, S.J.: Variational particle approximations. {\em The Journal of Machine Learning Research}, {\bf 18}, 2328-2356 (2017).



\bibitem{tarantola05}
Tarantola, A.: {\em Inverse problem theory and methods for model parameter estimation} (Vol. 89). {SIAM} (2005).

\bibitem{vapnik13}
Vapnik, V.: {\em The nature of statistical learning theory.} Springer science \& Business Media (2013).


\bibitem{zhou08}
  Zhou, D.X.: Derivative reproducing properties for kernel methods in learning theory. {\em Journal of computational and Applied Mathematics}, {\bf 220}, 456-463 (2008).
}
\end{thebibliography}
\end{document}